\documentclass[runningheads]{llncs}
\usepackage[T1]{fontenc}

\usepackage{microtype}
\usepackage{hyperref}
\usepackage{url}
\usepackage{booktabs}
\usepackage{amsmath}
\usepackage{graphicx}
\usepackage{hyperref}

\usepackage{graphicx}
\begin{document}
\title{S$^3$: A Simple Strong Sample-effective \\ Multimodal Dialog System}
%\author{Elisei Rykov\inst{1}\orcidID{0000-0002-8939-619X} \and
%Egor Malkershin\inst{1}\orcidID{0009-0008-3623-4040} \and
%Alexander Panchenko\inst{1,2}\orcidID{0000-0001-6097-6118}}
\author{Elisei Rykov\inst{1} \and
Egor Malkershin\inst{1} \and
Alexander Panchenko\inst{1,2}}
\authorrunning{E. Rykov et al.}

% \institute{Skolkovo Institute of Science and Technology}

\institute{Skolkovo Institute of Science and Technology, Russia \and Artificial Intelligence Research Institute, Russia\\
\email{ \href{mailto:e.rykov@skol.tech}{\{e.rykov, egor.malkershin, a.panchenko\}@skol.tech}}}
\maketitle              % typeset the header of the contribution
\begin{abstract}
In this work, we present a conceptually simple yet powerful baseline for multimodal dialog task, an S$^3$ model, that achieves near state-of-the-art results on two compelling leaderboards: MMMU and AI Journey Contest 2023. The system is based on a pre-trained large language model, pre-trained modality encoders for image and audio, and a trainable modality projector. The proposed effective data mixture for training such an architecture demonstrates that a multimodal model based on a strong language model and trained on a small amount of multimodal data can perform efficiently in the task of multimodal dialog.

% The observed approach requires a relatively small amount of training data and computational resources while demonstrating scores comparable to those of state-of-the-art models across various datasets. 
\end{abstract}

\section{Introduction}
In the dynamic landscape of artificial intelligence (AI), the advent of multimodal systems has marked a transformative shift, enabling machines to interpret and analyze heterogeneous data streams with unprecedented finesse. These systems, which seamlessly integrate multiple forms of data such as text, images, and audio, are becoming progressively adept at mirroring human cognitive capabilities. However, one of the principal challenges confronting researchers in this domain has been the necessity for considerable volumes of data and substantial computational resources to train state-of-the-art models.

\begin{figure}
    \centering
\resizebox{10cm}{!}{
    \includegraphics{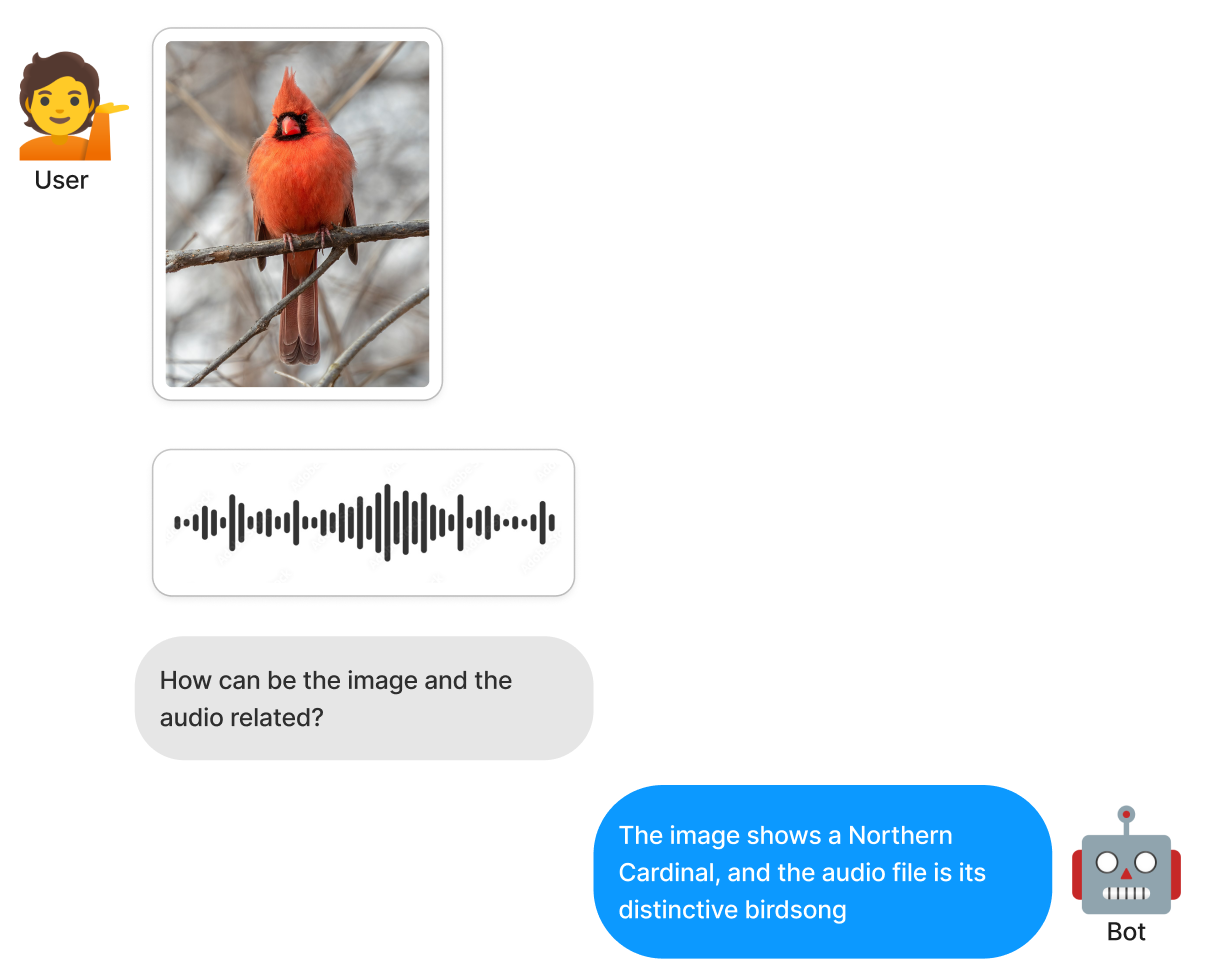}
}
    \caption{An example of multimodal dialog.}
    \label{fig:multimodal_dialog}
\end{figure}

Against this backdrop, our study introduces a novel paradigm that posits that a powerful multimodal system is feasible with minimal data and computational resources. This paper presents a simple yet effective baseline model which challenges the conventional premise that large datasets and excessive computational power are prerequisites for developing competitive multimodal AI systems. By using a compact corpus of less than 150,000 multimodal samples, a pre-trained frozen modality encoder, a 7B language model, and by exploiting the computational economy of a single A100-80GB GPU, we have created a model with an elegantly simple architecture that delivers performance on par with the more complex systems that currently dominate the field. The core of our approach is a modality projector that uses a simple multi-layer perceptron (MLP) to map multimodal features into token embeddings.

Our contributions can be summarized as follows:
\begin{itemize}
    \item We apply a well-known pipeline for training multimodal projectors to multiple modalities (image, audio, and text) to train a multimodal dialog model.
    \item We introduce a high-quality effective data mixture for training multimodal dialog models.
    \item We demonstrate that mapping of the whole image into 4 textual tokens is enough for multimodal dialog task.
    \item We openly release the obtained model, which shows comparable performance to state-of-the-art models.\footnote{\url{https://github.com/s-nlp/s3}}
\end{itemize}

\section{Related work}
In this section we consider relevant research on deep pre-trained models of various modalities (text, image, audio), the related multimodal dataset used for training such models, and the existing multimodal architectures based on both multimodal encoders and language models.

\subsection{Text Modality: Large Language Models (LLMs)}

% As for LLMs, there are two models that got our attention: LLaMA \cite{touvron2023llama} and Mistral \cite{jiang2023mistral}. Both of them were state of the art models in a recent time. LLaMA has been trained on a mixed dataset from various sources covering diverse domains. In comparison to LLaMA, Mistral introduces some changes. Firstly, Sliding Window Attention (SWA) to leverage stacked layers of a transformer for attending information beyond a window size W. The attention span is theoretically extended to approximately 131K tokens by using a window size of W = 4096 at the last layer. Notably, modifications inspired by FlashAttention and xFormers result in a 2x speed improvement over a vanilla attention baseline for a sequence length of 16K. The Rolling Buffer Cache efficiently manages a fixed cache size of W, reducing memory usage by 8x on a sequence length of 32K tokens without compromising model quality. Additionally, the approach incorporates Pre-fill and Chunking strategies, enabling the pre-filling of the cache with known prompts and chunking large prompts for efficient attention computation.

State-of-the-art large language models are based on deep pre-trained transformers. Popular representatives of this kind are LLaMA \cite{touvron2023llama} and Mistral \cite{jiang2023mistral}, which are used in our work. LLaMA was trained on a mixed dataset from different sources covering different domains. Compared to LLaMA, Mistral introduces some changes related to sliding window attention, pre-filling and chunking strategies, etc. 
%Firstly, Sliding Window Attention (SWA) to leverage stacked layers of a transformer for attending information beyond a window size W. The attention span is theoretically extended to approximately 131K tokens by using a window size of W = 4096 at the last layer. Notably, modifications inspired by FlashAttention and xFormers result in a 2x speed improvement over a vanilla attention baseline for a sequence length of 16K. The Rolling Buffer Cache efficiently manages a fixed cache size of W, reducing memory usage by 8x on a sequence length of 32K tokens without compromising model quality. Additionally, the approach incorporates Pre-fill and Chunking strategies, enabling the pre-filling of the cache with known prompts and chunking large prompts for efficient attention computation.

\subsection{Image Modality}

ImageBind \cite{girdhar2023imagebind} serves as a universal multimodal encoder, using a multi-layer architecture that facilitates the extraction of universal image features. One of the modern neural network encoders is CLIP \cite{radford2021learning}, which is based on a combined architecture of a convolutional neural network and a transformer. CLIP is trained on large datasets of images and text and is involved in contrastive tasks. The CLIP architecture enables the linking of visual and textual representations, providing an effective representation of both content and meaning. The effectiveness of CLIP allows the model to successfully classify distorted or context-dependent images and to integrate different knowledge domains into a unified space.

\subsection{Audio Modality}
Whisper \cite{radford2023robust}, an audio encoder, has received considerable attention in recent studies as an innovative encoder architecture. Unlike traditional encoders, Whisper is designed to produce compact and high-quality representations of input data. The architecture uses stacked autoencoders and introduces a pioneering training approach known as "WhisperNet". In addition, Imagebind can also be used to provide audio features.

% The training of CLIP involves the use of masked language models and a loss function based on vector products. The loss function is formulated as follows:

% \[ L(\theta) = \sum_{i=1}^{N} \max(0, \alpha - \text{sim}(f(x_i), g(y_i)) + \text{sim}(f(x_i), g(y_j))) \]

% where \(N\) is the number of training samples, \(\alpha\) is a margin hyperparameter, \(f(x_i)\) represents the visual representation of the \(i\)-th image, \(g(y_i)\) and \(g(y_j)\) are the textual representations of the \(i\)-th and \(j\)-th captions, and \(\text{sim}(\cdot)\) denotes the cosine similarity.
% This formulation encourages the positive pair (matching image and text) to have a higher similarity score than the negative pair (mismatched text), with the margin \(\alpha\) ensuring a separation between them.

\subsection{Datasets}
% An important component is also the selection of datasets. 
In the context of image captioning, datasets such as COCO \cite{lin2015microsoft} and TextCaps \cite{sidorov2020textcaps} have been created. The COCO dataset was created by collecting a large number of images with corresponding captions from the web, creating a diverse data set for training and evaluating image captioning models. The TextCaps dataset was created by extending the COCO dataset with additional descriptions obtained from contextual user queries. 

% This dataset presents unique challenges for "image captioning with reading comprehension," including determining relationships between OCR tokens and visual context, switching between model vocabulary and OCR tokens during caption generation, paraphrasing and making inferences about OCR tokens, and handling OCR tokens, including those unseen before (zero-shot). These tasks underscore the complexity of understanding both textual and visual elements in image captioning systems.

For the Visual Question Answering (VQA) task, VisDial \cite{das2017visual} was created using images and a series of dialogs about them to train models to answer questions about the images provided. ScienceQA \cite{lu2022learn} was created by automatically extracting questions and answers from textbooks and tests on science topics. VizWiz \cite{gurari2018vizwiz} was created using a mobile application that allows blind and visually impaired people to ask questions about unfamiliar images. Visual Genome is a dataset of images, attributes and relationships between objects in those images, created with the help of annotators. The GQA \cite{ainslie2023gqa} dataset was created using a powerful question engine based on the scene graph structures of Visual Genome, generating 22 million different questions about the visual world, with functional programs for answer control and bias smoothing. LLaVA dataset \cite{liu2023visual} is automatically compiled using GPT-4 with various prompts, including regular VQA dialogs, highly complex questions and more. For the optical character recognition (OCR) task, OCR\_books was used. OCR\_books contains images of text obtained after scanning books of various genres.

For the task of audio captioning, one of the well-known datasets is CLOTHO \cite{drossos2019clotho}, a fully crowdsourced dataset with multiple captions for each audio file.

The OpenAssistant dataset \cite{köpf2023openassistant} is a large collection of conversational data collected through crowdsourcing with over 13,000 volunteers. 

%Data collection involved a web-app interface and five steps: prompting, labeling prompts, adding reply messages, labeling replies, and ranking assistant replies. Volunteers were fully briefed on contributing to a public dataset, and content moderation and spam filtering ensured high quality and safety. With over 625,000 tasks completed, the dataset comprises over 10,000 fully annotated Conversation Trees. Appendices C and G offer UI displays and collection parameter settings respectively

\subsection{Multimodal Models}
% In this section will be discussed the existing solutions on the Multimodal LLM task. FROMAGe \cite{koh2023grounding} is the approach underlying our solution. The FROMAGe architecture is based on the idea of combining a pre-trained frozen LLM ($\rho_{\theta}$, where $\theta$ represents the non-trainable weights of the language model) with an equally pre-trained and frozen visual image encoder ($v_{\phi}$, where $\phi$ represents the non-trainable weights of the visual encoder) through just one layer of linear projection ($W_{c}R_{m} \times kd$, where $m$ is the dimensionality of visual embeddings, $k$ is the number of embeddings after projection, and $d$ is the dimensionality of the language model). This linear layer, although having a small number of trainable parameters, plays a crucial role in establishing the connection between the image and text modalities. For LLaVA \cite{liu2023visual} the authors employed GPT-4 to create a multimodal visual-text instructional dataset for training. LLaVA combines a LLM Vicuna and a visual encoder based on ViT-L/14 from CLIP, linked by a linear projection layer.

% In this section will be discussed the existing solutions on the Multimodal LLM task. 
The FROMAGe \cite{koh2023grounding} architecture is based on the idea of combining a pre-trained frozen LLM with an equally pre-trained and frozen visual image encoder through just one layer of linear projection. This linear layer, although having a small number of trainable parameters, plays a crucial role in establishing the connection between the image and text modalities. For LLaVA model \cite{liu2023visual} the authors employed GPT-4 to create a multimodal visual-text instructional dataset for training. LLaVA combines a LLM Vicuna and a visual encoder based on ViT-L/14 from CLIP, linked by a linear projection layer. Qwen-VL \cite{qwen-vl} demonstrates a similar approach to LLaVA with fine-tuning of the projector and multi-level training. HoneyBee \cite{honeybee} shows that classical approaches to building projectors from LLaVA could be improved by applying complex and effective resempling to reduce the number of modality tokens. CogVLM \cite{cogvlm} introduces a so-called "visual expert" module: a copy of some transformer blocks inside the language model that activate and update their weights only on image tokens.

\section{Methods}

\begin{figure}
    \centering
\resizebox{13cm}{!}{
    \includegraphics{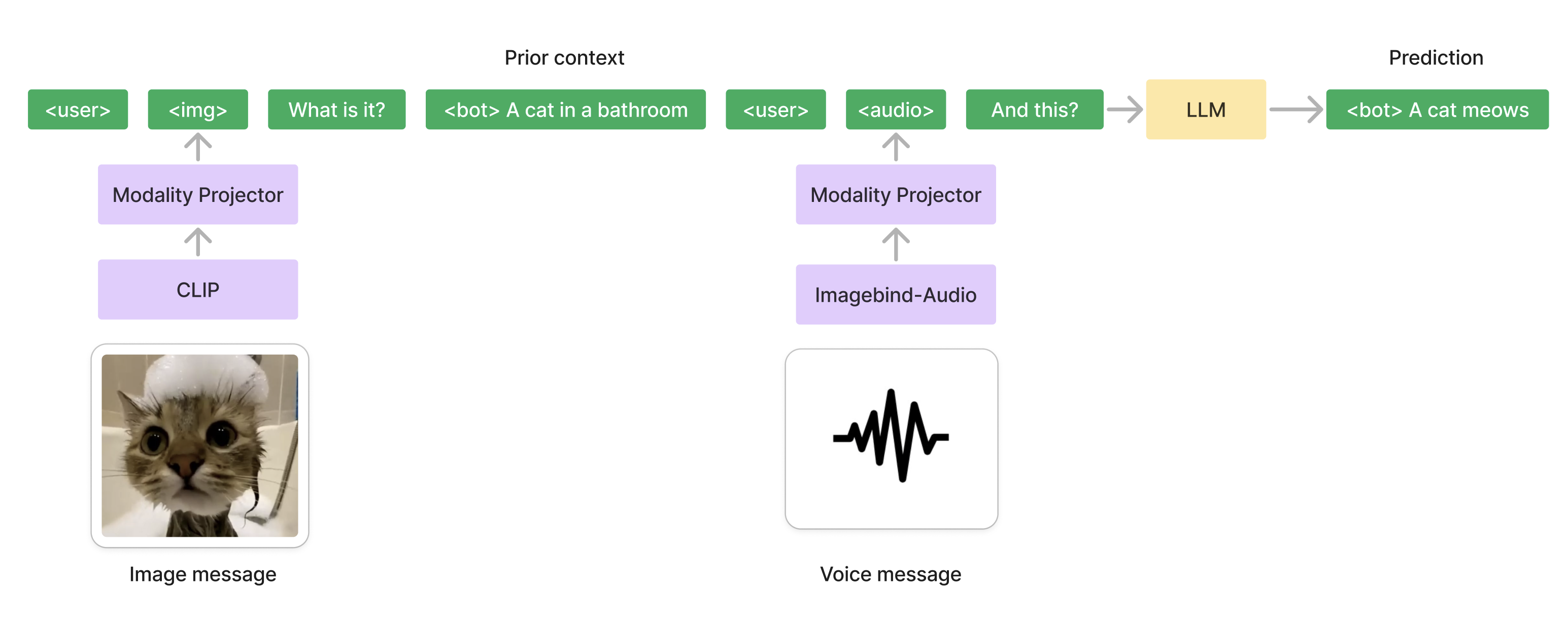}
}
    \caption{Architecture of S$^3$ multimodal dialog system. All modalities are passed to specific encoders, and then modality objects are passed to modality projectors, which map them to token embeddings of a Large Language Model.}
    \label{fig:architecture}
\end{figure}

\subsection{Dataset preprocessing}
We developed a model designed to engage in multimodal conversations with users. To achieve this, we formatted each dataset in a standard chat layout. This format involves representing each message as a JSON object containing 'role' (indicating whether the message is from a user or the bot), 'type' (indicating whether the message contains an image, audio, or text), and the message content itself (this would be the file path in case of images and audios). Our chat layout allows media content to be inserted at any point in the conversation, possibly multiple times.

\begin{figure}[h]
    \centering
    \begin{footnotesize}
    \begin{verbatim}
{
  "id": 0,
  "messages": [
    {
      "role": "user",
      "type": "image",
      "text": "https://example.com/images/bird.jpg"
    },
    {
      "role": "user",
      "type": "audio",
      "text": "https://example.com/audio/birdsong.mp3"
    },
    {
      "role": "user",
      "type": "text",
      "text": "How can be the image and the audio related?"
    },
    {
      "role": "bot",
      "type": "text",
      "text": "The image shows a Northern Cardinal, 
               and the audio file is its distinctive birdsong."
    }
  ]
}
    \end{verbatim}
    \end{footnotesize}
    \caption{Example of json-formatted multimodal dialog data.}
    \label{fig:enter-label}
\end{figure}

For each dataset, we created a custom system prompt that was tailored to elicit bot responses that closely matched the original dataset. For example, for the TextCaps dataset, we chose a prompt such as ``Answer the question with a single word or phrase'' to reflect the fact that the dataset contains primarily short responses. By implementing such prompts, we can guide the model to respond either concisely or with elaborate explanations, depending on the context. When adapting captioning datasets such as COCO or CLOTHO-Captions for conversational use, we formulated a set of basic synthetic questions, such as ``What do you see in this picture?'' or ``What could make this sound?''. These artificial questions serve as prompts to the user about a particular image or sound. 

In our setup, the appearance of an image within a conversation is flexible. We randomised the order of questions and corresponding images within each dataset, allowing the media content to precede or follow the question. This randomness addresses the issue of brevity that is present in many datasets, which typically consist of single pairs of questions and answers. To create more extended dialogs and overcome this limitation, we randomly combined several short dialogs into extended sequences. 

We introduced a unique mixture of data specifically designed for our task, as detailed in Table \ref{tab:mixture}. Our goal was to create a diverse collection, for which we included a range of tasks such as Optical Character Recognition (OCR), Visual Question Answering (VQA), Audio Question Captioning (AQA), image captioning, casual conversation, and more. In total, we trained the model on around 145,000 samples.

\begin{table*}[]
\resizebox{\textwidth}{!}{
\begin{tabular}{@{}llll@{}}
\toprule
\textbf{Task}  & \textbf{Dataset} & \textbf{Samples in train/val} & \textbf{System prompt}                                          \\ \midrule
Image          & COCO             & 5000                          & Provide a one-sentence answer for the provided question         \\
               & GQA              & 10000                         & Answer the question using a single phrase                       \\
               & ImageChat        & 5000                          & Show reaction and emotion in response to images                 \\
               & Visual Dialog    & 24000                         & Answer the question using a single word or phrase               \\
               & LLaVA            & 31000                         & Answer questions thoroughly and in detail                       \\
               & ScienceQA        & 10000                         & Answer with the option’s letter from the given choices directly \\
               & OCR-VQA          & 1000                          & Answer the question using a single word or phrase               \\
               & TextCaps         & 14000                         & Answer the question using a single word or phrase               \\
               & VizWiz           & 10000                         & Answer the question using a single word or phrase               \\
               & Visual Genome    & 5500                          & Answer the question using a single word or phrase               \\
               & OKVQA            & 5000                          & Answer the question using a single word or phrase               \\
               & AOKVQA           & 10000                         & Answer questions thoroughly and in detail                       \\
Audio          & CLOTHO-Captions  & 3750                          & Answer the question using a single word or phrase               \\
               & CLOTHO-AQA       & 1000                          & Answer the question using a single word or phrase               \\
               & AudioSet         & 5000                          & Provide a one-sentence answer for the provided question         \\
Text           & OpenAssistant    & 5000                          & You are helpful AI assistant                                    \\ \midrule
\textbf{Total} & -                & 145250                        & -                                                               \\ \bottomrule
\end{tabular}
}
\caption{Training data mixture used to train S$^3$.}\label{tab:mixture}
\end{table*}

\subsection{Special tokens and post processing}
We integrated additional special tokens into the tokenizer of the base model and unfrozen both the language model head and the embedding layer to facilitate the training of these new tokens. Specifically, we introduced modality tokens \texttt{[M]} and \texttt{[/M]} to mark the beginning and end of different modality objects in the data. We encoded different modality objects using tokens such as \texttt{[audio]} and \texttt{[img]} for audio and image content, respectively. To represent speakers within the dialogs, we included special tokens for the bot and user roles. We also included the \texttt{[RS]} token to indicate the start and end of each message in a dialog. The \texttt{[M]} tokens were specifically used to indicate the scope of modality objects. Consequently, each dialog in the dataset is processed into a string containing these special tokens, as shown below:
\begin{verbatim}
[RS][user][M][img][img][img][img][/M][/RS]
[RS][user]What is it?[/RS]
[RS][bot]A red apple with red worm[/RS]
\end{verbatim}

In the data processing stage, an image processor handles all the images, which are then transformed into image embeddings via an encoder. Our multimodal dialog model is designed to accept embeddings as the primary form of input data. Therefore, in the preprocessing phase, we tokenize the processed chat data and retrieve the embeddings for each token. For multimodal tokens such as \texttt{[img]} and \texttt{[audio]}, we replace them with the output embeddings of the corresponding image, which are divided into N segments. In our configuration, we chose to split the modality embeddings into four different tokens.

\subsection{Model architecture}
We used a widely accepted a ``shallow alignment'' architecture, which consists of three main components: a basic Large Language Model, a modality encoder, and a modality projector. The role of the modality encoder is to generate image representations, which are then transformed into token embeddings by the modality projector, allowing the integration of visual information into the language model. The architecture of the whole multimodal dialog model is shown in figure \ref{fig:architecture}.

\subsection{Pre-trained modality encoder}
Modality encoders are designed to preprocess different types of modality objects, such as images and audio, and transform them into embeddings. These embeddings are then made compatible with the Large Language Model by a modality projector. To process audio inputs, we experimented with the ImageBind multimodal encoder, which can handle various modalities, including audio, image, and video. For image inputs, we used CLIP as the encoder of choice. Typically, image encoders such as CLIP aggregate the output of the processed modality object, usually by pooling, to produce a single, final embedding. However, for the purposes of our task, we hypothesized that using individual patched embeddings would yield more advantageous results.

\begin{figure}
    \centering
\resizebox{10cm}{!}{
    \includegraphics{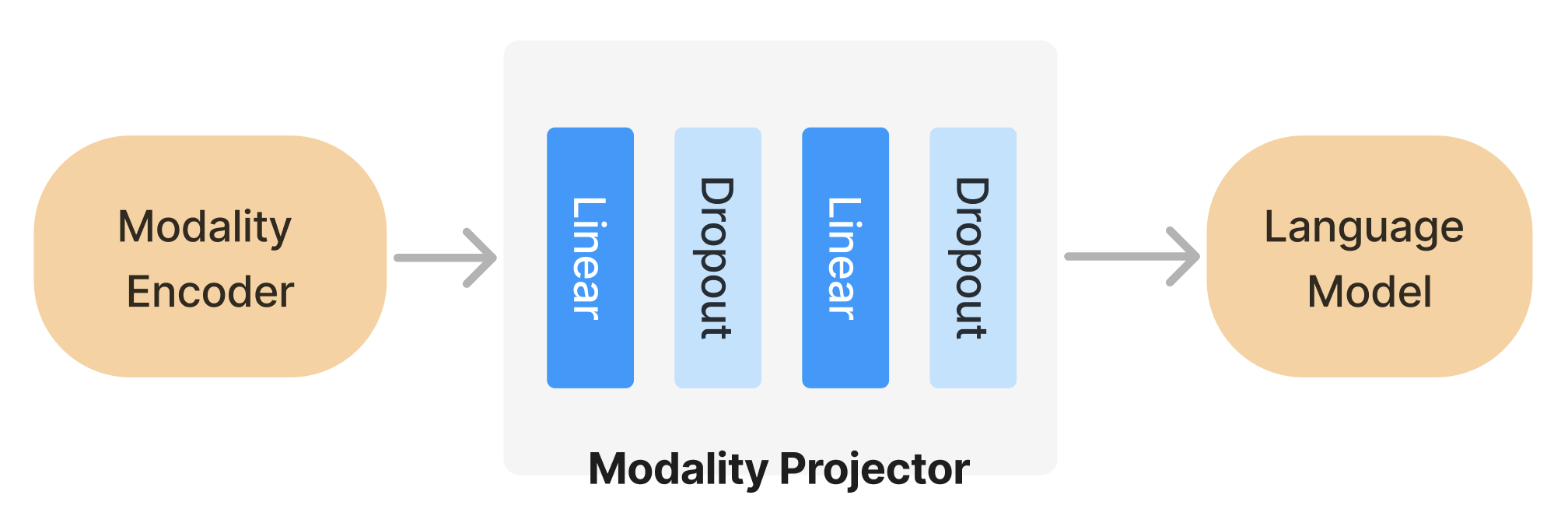}
}
    \caption{Architecture of our MLP modality projector, which maps features from the modality encoder to the language model.}
    \label{fig:projector}
\end{figure}

\subsection{Modality projector}
The role of the modality projector is to adjust the embeddings of various modality objects, such as images and audio, to ensure they are compatible with the language model. Its configuration can vary from a straightforward linear layer to a more intricate Multilayer Perceptron, among other options. In our research, we initially implemented a basic architectural design in which the hidden states from the modality encoder are mapped directly to the token embeddings of the language model using several linear layers. The complete architecture of our MLP modality projector is shown in figure \ref{fig:projector}. This modality projector is capable of converting an image only into a 4 token embeddings by extending the output from its final linear layer. To illustrate, if the dimension of the token embeddings within the language model is 32, we can instruct the output projection of the modality projector to produce a 128-dimensional embedding, which we then divide into 4 separate token embeddings. We employed the same architectural design for the modality projector across both image and speech modalities. In contrast to state-of-the-art models such as LLaVA, we map the modality object into 4 tokens, regardless of the number of output patches within the modality encoder. We assume that the small number of output modality tokens is sufficient for basic visual understanding. Also, using only 4 tokens significantly reduces the length of the sequence we pass to Transformer.

\subsection{Language model}
We incorporated parameter-efficient fine-tuning LoRA adapters into our baseline language model to enhance its interaction capabilities in dialog contexts. Starting with basic pre-trained models such as LLaMA or Mistral, which do not inherently have the ability to participate in user dialogs, we postulated that the integration of a PEFT adapter would solve this problem. In the course of our experiments, we found that the Mistral-7b model served as an effective base model. By using PEFT adapters, we aimed not only to provide the language model with improved conversational capabilities, but also to do so without significantly increasing the number of trainable parameters; this efficient approach leverages the existing knowledge of the pre-trained model and complements it with targeted tunability for specific dialog-oriented tasks.

\subsection{Training details}
The entire architecture was trained using the CrossEntropyLoss function. To speed up the training process, we used DeepSpeed, especially at optimization level 2, which is designed to provide significant speedups by introducing optimizations that reduce memory usage and improve computational performance. For the training procedure, we created a custom training pipeline rooted in the HuggingFace trainer framework. Within this configuration, we used the AdamW optimizer. To regulate the learning rate, we used a cosine annealing scheduler, starting with an initial learning rate of 1e-4. The batch size was set to 128. All training was done on a single A100-80GB GPU.

\section{Results}
In this section, we thoughly test our model in two recent compeltive leaderboards namely AI Journey Content 2023\footnote{\url{https://dsworks.ru/en/champ/super-aintelligence\#overview}} and MMMU~\cite{yue2023mmmu}

\subsection{Experiment 1: AI Journey}
The first trial of our multimodal framework was tailored for the AI Journey Contest 2023. AI Jorney Contest is an annual competition related to ML and AI with cash prizes from large companies, an analog of the Kaggle platform. 

Our system was developed to solve the ``Strong Intelligence'' task in the AI Jorney Contest 2023. The goal of this contest was to develop a system with the ability to interact seamlessly across three modalities: text, image, and audio. For reasoning, we used the instruction ``Answer questions thoroughly and in detail'', which was given to encourage the model to provide comprehensive answers with ample elaboration. Thus, an effective system should demonstrate adeptness in managing input across images, audio, and text.

During the evaluation, all systems were constrained by the following inference container conditions: 243 GB RAM, 16 CPU cores, 1 Tesla A100 (80 GB) GPU, 3.5 hours to run, and no access to Internet resources. 
% \begin{itemize}
%     \item 243 Gb RAM
%     \item 16 CPU-cores
%     \item 1 GPU Tesla A100 (80Gb)
%     \item 3.5 hours for execution
%     \item No access to internet resources
% \end{itemize}

\textbf{dataset}: The evaluation phase challenged the system by presenting dialogs of heterogeneous composition: some consisted of only textual prompts, some combined text and images, some intertwined text and audio, and some even contained all three modalities. The dialogs presented to the system could involve contexts spanning more than two exchanges, allowing for the possibility of recurrence of image modality objects within a single dialog. Furthermore, in scenarios involving multiple modalities, there was potential for intermodality relationships, such as a user asking the bot to identify differences between two provided images. A notable detail of the contest setup was the absence of a specifically designated evaluation dataset, as provided by the contest organizers. Because the evaluation process was private, the evaluation data and the exact description of the other teams' submissions are not available. 

\textbf{Metric:} The evaluation of our multimodal system was based on two metrics: METEOR \cite{banerjee-lavie-2005-meteor} and a Hidden Metric based on the perplexity of the model's answer.

\begin{center}
\begin{math}
    HM(X) = exp(\frac{1}{t}\sum_{i}^{t}\log{p_{\theta}(x_{i}|x_{<i})}),
\end{math}
\end{center}

where $ \log{p_{\theta}(x_{i}|x_{<i}}) $ is the likelihood of the $i$th token given all $x < i$ tokens. The final Integral Metric:

\begin{center}
\begin{math}
    \sum_{j=1}^{J}\frac{\omega_{j}}{N_{j}}\sum_{i=1}^{N_{j}} \frac{METEOR_{i} + HM_i}{2},
\end{math}
\end{center}

where $j$ is type of dialog, $N_j$ is the number of dialogs for of type $j$, and $\omega_j$ is the weight of examples for the dialogs of the $j$ type. For textual only dialogs, authors propose $\omega_j = 0.1$. For dialogs with both text and images, $\omega_j = 0.2$. For dialogs with audios and text, $\omega_j = 0.3$. And, finally, for dialogs that consists of all three modalities, $\omega_j = 0.4$.

\textbf{Result:} In the AI Journey contest, our approach secured 4th place out of a total of 30 participating teams. The top 10 systems from the contest, along with detailed metrics, are presented in the table \ref{tab:ai-jorney}. The best-performing approach demonstrated, similar to our system setup, additionally included a small sample of self-generated supervised high-quality training data for the LoRA fine-tuning phase. Therefore, the author of the system freezes the language model at each stage of training, except the last one, using self-generated data.

\begin{table}
\begin{center}
\setlength\tabcolsep{0.35cm}
\begin{tabular}{@{}lllll@{}}
\toprule
\textbf{Rank} & \textbf{Team name} & \textbf{HM} & \textbf{METEOR} & \textbf{Average}   \\ \midrule
1             & fffrrtt            & 0.576               & 0.522                & 0.549          \\
2             & gradient sunset    & 0.534               & 0.455                & 0.495          \\
3             & whatever           & 0.517               & 0.458                & 0.487          \\
\textbf{4}    & \textbf{S$^3$ (Ours)}      & {0.549}      & {0.414}       & {0.481} \\
5             & DeReyly            & 0.498               & 0.427                & 0.462          \\
6             & Baseline            & 0.497               & 0.370                & 0.434          \\
7             & cybercho            & 0.489               & 0.377                & 0.433          \\
8             & EvilAI            & 0.472               & 0.395                & 0.433          \\
9             & AgaUgu            & 0.486               & 0.359                & 0.423          \\
10            & Denisiuskley            & 0.491               & 0.337                & 0.414          \\ \bottomrule

\end{tabular}
\end{center}
\caption{Official rankings of top-10 submits out of 30 teams in AI Journey Contest of multimodal dialog systems.}
\label{tab:ai-jorney}
\end{table}

\begin{figure}
    \centering
\resizebox{\textwidth}{!}{
    \includegraphics{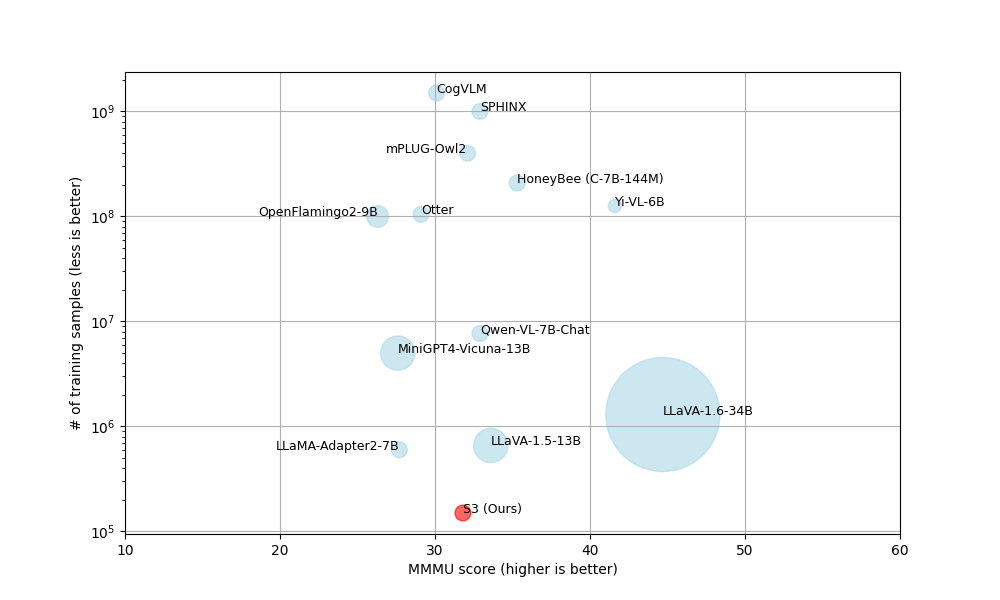}
}
    \caption{A comparative analysis of the performance of S$^3$. According to the MMMU benchmark score, S$^3$ shows a competitive score in comparison to various models with a larger size and more training samples. The size of marks corresponds to the number of parameters in the language model.}
    \label{fig:scatter}
\end{figure}

\subsection{Experiment 2: MMMU}
The model we developed was then evaluated on the MMMU benchmark to determine its visual comprehension abilities and to compare it to existing models. When generating responses for the MMMU dataset, we instructed the model using the prompt ``Answer with the letter of the option directly from the given choices'' to ensure that the responses were concise and directly related to the given answer choices.

\textbf{Dataset:}
The MMMU benchmark serves as a comprehensive resource for evaluating the capabilities of multimodal dialog models. It includes a dataset of over 11,500 questions curated from college-level exams, quizzes, and textbooks spanning six academic disciplines: arts, business, science, health, humanities, and engineering. In total, the questions cover 30 subjects and delve into 183 specific subfields, incorporating a wide range of 30 different image types, including but not limited to graphs, diagrams, and chemical structures.

\textbf{Metric:} The performance on the MMMU benchmark is quantified by calculating the average accuracy across the different types of tasks included in the benchmark.

\textbf{Result:} Our system, which utilizes a low-count data mixture, surpassed many of the existing models in terms of performance. Across open-sourced 7B models, it demonstrates a competitive performance. It even closely competed with state-of-the-art models that were trained on significantly larger datasets, with larger LLM, falling short by a marginal difference. Our model's standing, along with comparisons to other models, can be found in a segment of the MMMU leaderboard that is shown in Table \ref{tab:mmmu}.

\begin{table}[]
\resizebox{\textwidth}{!}{
	\begin{tabular}{l|p{2.3cm}|p{2.3cm}|p{2.3cm}|p{2.3cm}|p{2.3cm}}
		\toprule
\textbf{Model} & \textbf{Model is open sourced} & \textbf{Leaderboard test score} & \textbf{\# of params of LLM} & \textbf{\# of params visual encoder} & \textbf{\# of training data samples} \\ \midrule
		GPT-4V(ision) (Playground) & -- & 55.7                & --                            &                                      & --                            \\
		Qwen-VL-MAX     \cite{qwen-vl}                                                           & --                     & 46.8                & 7.7B                         & 1.9B                                  & --                            \\
		LLaVA-1.6-34B     \cite{liu2023improvedllava}                                            & +                     & 44.7                & 34B                          & 0.4B                                  & 1.3M                         \\
		Yi-VL-6B         \cite{ai2024yi}                                                         & +                     & 41.6                & 6B                           & 1B                                    & 126M                         \\
		HoneyBee (C-7B-144M)  \cite{honeybee}                                                  & +               & 35.3                & 7B                        & 0.4B                                  & 208M             \\
  BLIP-2 FLAN-T5-XXL  \footnote{\url{https://huggingface.co/Salesforce/blip2-flan-t5-xxl}} & +                     & 34.0                & 11.3B                        & 0.4B                                  & --                            \\
		InstructBLIP-T5-XXL   \cite{NEURIPS2023_9a6a435e}                                        & +                     & 33.8                & 11.3B                        & 0.4B                                  & --                            \\
		LLaVA-1.5-13B    \cite{liu2023improvedllava}                                             & +                     & 33.6                & 13B                          & 0.4B                                  & 0.66M                        \\
		Qwen-VL-7B-Chat    \cite{qwen-vl}                                                        & +                     & 32.9                & 7B                           & 1.9B                                  & 76.8M                        \\
  		SPHINX    \cite{sphinx}                                                        & +                     & 32.9                & 7B                           & 0.4B                                  & 1000M                        \\
		mPLUG-Owl2    \cite{m-plug}                                                        & +                     & 32.1                & 7B                           & 0.4B                                  & 400M                        \\

		\textbf{S$^3$ (Ours)}                                                                            & +                     & \textbf{31.8}       & \textbf{7B}                  & \textbf{0.4B}                         & \textbf{0.15M}               \\
		CogVLM       \cite{cogvlm}                                    & +                     & 30.1                & 7B                           & 0.3B                                  & 1500M                        \\
		Otter    \cite{otter}                                        & +                     & 29.1                & 7B                           & 0.4B                                  & 105M                         \\
		LLaMA-Adapter2-7B     \cite{DBLP:journals/corr/abs-2304-15010}                           & +                     & 27.7                & 7B                           & --                                     & 0.6M                         \\
		MiniGPT4-Vicuna--13B    \cite{DBLP:journals/corr/abs-2304-10592}                          & +                     & 27.6                & 13B                          & --                                     & 5M                            \\
		Adept Fuyu-8B     \footnote{\url{https://huggingface.co/adept/fuyu-8b}}                  & +                     & 27.4                & 8B                           & --                                     & --                            \\
		Kosmos2       \cite{DBLP:journals/corr/abs-2306-14824}                                   & +                     & 26.6                & --                            & --                                     & 90M                          \\
		OpenFlamingo2-9B     \cite{DBLP:journals/corr/abs-2308-01390}                            & +                     & 26.3                & 9B                           & 0.4B                                  & 100M                            \\
		Frequent Choice                                                                          & --                     & 23.9                & --                            & --                                     & --                            \\
		Random Choice                                                                            & --                     & 25.8                & --                            & --                                     & --                            \\ \bottomrule
	\end{tabular}
}
\caption{The MMMU leaderboard as compared to our  multimodal dialog system.}
\label{tab:mmmu}
\end{table}

\subsection{Conclusion}
% Moreover, our model's proficiency, as evidenced by its performance on the prestigious AI Journey contest and the MMMU benchmark, underlines the potential of simpler, efficient designs to meet or even surpass the benchmarks set by more resource-intensive models. Particularly in scenarios where data is scarce or computational resources are limited, this model offers a promising alternative. The versatility of our approach also extends to broader applications across different fields and disciplines, as evidenced by the diversity of subjects covered in the MMMU benchmark, ranging from art to engineering. By incorporating domain-specific knowledge within a unifying framework, the model demonstrates an encouraging capacity to handle expert-level tasks with noteworthy precision. 

Our research demonstrates that it is possible to develop a highly competitive multimodal dialog model without the need for large datasets or enormous computing power. Using less than 150,000 multimodal samples and a single A100-80GB GPU, we constructed a system that performs comparably to state-of-the-art models in the field. In particular, our model features a simple architecture consisting of a modality projector that uses a simple multi-layer perceptron (MLP) to effectively integrate a substantial amount of information into token embeddings. 

Future work should focus on increasing dataset size and diversity, especially in the audio modality, as this could potentially lead to further performance gains. In addition, exploring the integration of more complex architectures for modality adaptation could also be beneficial for further enhancing its capabilities.

\section*{Acknowledgements}

The research of Alexander Panchenko was supported by the Russian Science Foundation grant 20-71-10135.
\bibliographystyle{splncs04}

\bibliography{s3}

\end{document}